%% file: main.tex
\documentclass[runningheads]{llncs}

\usepackage[english]{babel} 
\usepackage[utf8]{inputenc} 
\usepackage[T1]{fontenc}    

\usepackage{graphicx}       

\usepackage{amsmath}
\usepackage{amsfonts}
\usepackage{graphicx}
\usepackage{todonotes}
\usepackage{float}
\usepackage{listofitems}
\usepackage{tikz}
\usepackage{subcaption}
\usetikzlibrary{shapes,arrows,calc}

\usepackage[open,numbered]{bookmark} 

\usepackage[nocompress]{cite} 

\usepackage{hyperref}       

\usepackage[noabbrev,nameinlink]{cleveref}
\usepackage[acronym]{glossaries}

\begin{document}

\input{content/acronyms.tex}

\input{content/01-Title_Authors}

\input{content/00-Abstract}

\glsresetall

\input{content/01-Introduction}
\input{content/01.5-Application}
\input{content/02-Model_Description}
\input{content/03-Experiments}
\input{content/04-Conclusion}

\bibliographystyle{splncs04}
\bibliography{main}

\end{document}

%% file: content/acronyms.tex
\newacronym{ai}{AI}{artificial intelligence}
\newacronym{ann}{ANN}{artificial neural network}
\newacronym{ctn}{CTN}{continuous-time network}
\newacronym{dm}{DM}{direct manipulation}
\newacronym{dnn}{DNN}{deep neural network}
\newacronym{hci}{HCI}{human-computer interaction}
\newacronym{icf}{ICF}{\textbf{independently controllable factors of variation}}
\newacronym{ltc}{LTC}{\textbf{liquid time-constant network}}
\newacronym{ml}{ML}{machine learning}
\newacronym{mlp}{MLP}{multilayer perceptron}
\newacronym{mmi}{MMI}{multimodal, multisensor interfaces}
\newacronym{rl}{RL}{reinforcement learning}
\newacronym{xai}{XAI}{explainable AI}
\newacronym{xaines}{XAINES}{explaining AI with narratives}
\newacronym{sme}{SME}{small and medium-sized enterprise}
\newacronym{ssl}{SSL}{self-supervised learning}
\newacronym{vr}{VR}{virtual reality}

\newacronym{al}{AL}{active learning}
\newacronym{cst}{CST}{computational sustainability \& technology}
\newacronym{gui}{GUI}{graphical user interface}
\newacronym{ui}{UI}{user interface}
\newacronym{iml}{IML}{interactive machine learning}
\newacronym{pam}{PAM}{passive acoustic monitoring}

\newacronym{uav}{UAV}{underwater autonomous vehicle}

\newacronym{pca}{PCA}{principal components analysis}
\newacronym{vae}{VAE}{variational autoencoder}
\newacronym{tda}{TDA}{topological data analysis}

\newacronym{dgm}{DGM}{deep generative model}

%% file: content/01-Title_Authors.tex
\title{A Deep Generative Model for Interactive Data Annotation through Direct Manipulation in Latent Space}

\titlerunning{A Network Model for Interactive Latent Representation Learning}
%

\author{Hannes Kath\inst{1,2} \and
Thiago S. Gouvêa\inst{1} \and
Daniel Sonntag\inst{1,2}}

\authorrunning{H. Kath et al.}
%

\institute{
German Research Center for Artificial Intelligence (DFKI), Oldenburg, Germany \email{\{hannes\_berthold.kath, thiago.gouvea, daniel.sonntag\}@dfki.de}
\and
University of Oldenburg, Applied Artificial Intelligence (AAI), Oldenburg, Germany
}


\maketitle              

%% file: content/00-Abstract.tex
\begin{abstract}

The impact of  machine learning (ML) in many fields of application is constrained by lack of annotated data.
Among existing tools for ML-assisted data annotation, one little explored tool type relies on an analogy between the coordinates of a graphical user interface and the latent space of a neural network for interaction through direct manipulation.
In the present work, we
1) expand the paradigm by proposing two new analogies: time and force as reflecting iterations and gradients of network training;
2) propose a network model for learning a compact graphical representation of the data that takes into account both its internal structure and user provided annotations; and
3) investigate the impact of model hyperparameters on the learned graphical representations of the data, identifying candidate model variants for a future user study.

\keywords{Deep Generative Model \and Self-supervised Learning \and Variational Autoencoder}

\end{abstract}

%% file: content/01-Introduction.tex
\section{Introduction}
\label{sec:intro}

Modern \gls{ml} algorithms such as \glspl{dnn} have shown outstanding performance on various tasks, especially on supervised learning. 
Still, the impact of \gls{ml} in many application areas is limited by the low availability of annotated data, as well as by technical entry barriers for domain experts, leading to a cross-domain need for user-friendly tools to annotate vast datasets.
To address this need, we propose and investigate a \gls{dgm} for the development of a data annotation tool that visualises data samples in a 2D or 3D \gls{gui}. 

Previous work has shown that interactive \glspl{gui} using the latent space of \gls{ml} systems can facilitate data interaction \cite{prange2021demonstrator}. 
The authors show that \gls{ml} systems can learn representations of a data domain from which users can obtain actionable views using \glspl{gui}, such as \gls{vr}.
Another tool for \gls{ml}-assisted data annotation operating on latent representations is Seadash \cite{gouvea_annotating_2022}.
Seadash is a tool for graphical data programming \cite{ratner2016data} that operates based on redundant, parallel data transformations.

In this paper, we take a general approach to improving data efficiency, which is to first learn a representation that captures the data structure \cite{bengio2013representation}, then annotate some data points, and finally restructure the data incorporating new annotated data points, facilitating further annotation.
A powerful method for learning useful representations in an annotation-free manner is \gls{ssl} \cite{liu2021self,lecun_self-supervised_2021}.
In \gls{ssl}, a model learns to predict labels that are automatically computed from the input data itself, e.g. by reconstructing input. The learned representations are then fine tuned based on the class annotations provided by the user.
To exploit the limited amount of annotated data, we use a \gls{dgm}, following the promising approach of Kingma et al. \cite{kingma2014SSLwithDGM}. 
\gls{ssl} can be considered as a special task in generative models that involves imputing missing data as a means to improve classification accuracy \cite{kingma2014SSLwithDGM}.
\Glspl{dgm} learn entangled representations, which means that the dimensions of the latent space do not always have a consistent meaning along the axes \cite{paige2017learning}.
Using a \gls{dgm} that learns disentanglement \cite{paige2017learning} in such a way that the decision boundaries or the respective clusters are clearly visible to the user, we structure the latent space.

%% file: content/01.5-Application.tex
\section{Application}
\label{sec:application}

The \gls{dgm} described in this paper is designed to run a data annotation system with 2D or 3D representations. 
The overall system architecture is shown in \cref{fig:system}.
Leading to the model requirements outlined in \cref{sec:model}, this section provides a concise overview of the user's workflow.
Details about the development and presentation of the application can be found in separate publications for a 2D \gls{gui} \cite{kath2023PAM} and a 3D \gls{gui} \cite{kath2023vr}.

The workflow of the application follows the concepts of \gls{dm}, which focus on physical, incremental and reversible user interactions on displayed objects of interest whose effects are immediately visible \cite{Shneiderman97DirectManipulation}:
A partially labelled dataset is embedded in a latent 2D or 3D space represented in the \gls{gui}.
Similar inputs are clustered so that multiple data samples can be annotated at once (e.g. with the Lasso tool).
New annotated data samples are used to re-train the \gls{dgm}, which triggers movement of the data samples: data samples of the same class are pulled together while different classes are pushed apart, creating the illusion of an invisible force like magnetism.

By applying the concept of \gls{dm}, we derive three metaphors: space (latent space represented as interface space), time (intermediate training results represented as movement) and force (backpropagation represented as physical force).
In this work, we therefore focus on the learned embedding of data points, the training process and backpropagation of the \gls{dgm}, while the user interactions in terms of space, time and force take place in the \gls{gui}.

%% file: content/02-Model_Description.tex
\section{Model description}
\label{sec:model}

Functional requirements for the model derived from the system architecture in \cref{fig:system} include learning relevant data structures without annotations, mapping them into a 2D or 3D latent space, a way for the user to explore, inspect and annotate data, as well as a way to helpfully customize the latent space for the user based on the added annotations.
The central non-functional requirement derived from the metaphors described in \cref{sec:application} is that the user interface needs to reflect essential concepts of the \gls{ml} system, namely the latent space, the training progress and the backpropagation.
Considering these requirements, we developed a \gls{dgm} architecture (\cref{fig:model}) based on a \gls{vae} \cite{kingma2013auto} and extended it with a classification header.

\input{figures/model_scheme.tex}

\paragraph{\textbf{\Glsfirst{vae}.}}
\Cref{fig:model} shows the \glspl{vae} architecture besides the classification header.
The encoder $q_\phi(Z \mid X)$ captures data structures from the input data $X$ and represents them in the latent space using the representation $Z$.
The dimensionality of the latent space is chosen low
    to force the learned representation to capture the most salient features of the training data and
    to allow the representation of the entire state space in a 2D or 3D environment.
The decoder $p_\theta(X \mid Z)$ is used for training purposes and calculates the reconstruction $\Tilde{X}$ using $Z$.
The parameters $\phi$ and $\theta$ are learned jointly using gradient descent on random minibatches.
A \Gls{vae} is a probabilistic model, because each input $X$ induces a probability distribution over the entire latent space of variable $Z$.
In the standard implementation of a \gls{vae},
    $Z$ is parameterized as a multivariate Gaussian whose parameters $\mu_Z$ and $\Sigma_Z$ are computed deterministically from $X$, and downstream operations are performed on samples from this distribution \cite{kingma2013auto}.
While samples of $Z$ are used in learning, we propose to use the mean $\mu_Z$ for visualizations.

\paragraph{\textbf{Classification head.}}

The latent space of the \gls{vae} is displayed to the user for annotation.
A user might find it helpful that the space is transformed to group data points of the same category.
Since \glspl{vae} cannot handle annotations, we introduce the variable $Y$ and a classification head that computes the class labels $\Tilde{Y}$ from the latent variable $Z$.
This classification head is a simple \gls{mlp}.

\paragraph{\textbf{Loss functions.}}

The \gls{vae} and \gls{mlp} are jointly optimized by minimizing the sum of all loss components:
\begin{equation}
\label{eq:total_loss}
    \mathcal{L} = \mathcal{L}_{\text{reconst}}(X, \Tilde{X}) + \beta_{KL} D_{KL} \left( q_{\phi}(Z \mid X)  \mid\mid p(Z) \right) + \beta_{\text{classifier}} \mathcal{H}(Y, \Tilde{Y}),
\end{equation}
where the first two terms are from \cite{burgess2018understanding}, $\mathcal{H}$ is the cross entropy loss between $Y$ and $\Tilde{Y}$, and $\beta_{\text{KL}}$ and $\beta_{\text{classifier}}$ serve as the weighting coefficients for the loss terms.

%% file: figures/model_scheme.tex
\tikzstyle{mynode}=[thick,draw,circle,minimum size=12]
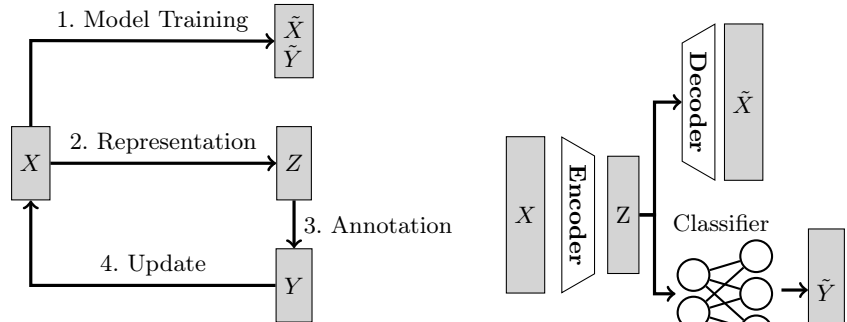
\begin{figure}[ht] 
     \centering 
     \begin{subfigure}{.45\textwidth} 
         \centering 
         \begin{tikzpicture} 
             \node[rectangle, draw, fill = {rgb:black,1;white,5}, minimum height=3em]
            (X) {$X$};
            \node[rectangle, draw, fill = {rgb:black,1;white,5}, minimum height=3em, right of = X, yshift = 5em, align = left, xshift=2.5cm]
            (XY) {$\Tilde{X}$ \\ $\Tilde{Y}$};
            \node[rectangle, draw, fill = {rgb:black,1;white,5}, minimum height=3em, right of = X, yshift = 0em, xshift=2.5cm]
            (Z) {$Z$};
            \node[rectangle, draw, fill = {rgb:black,1;white,5}, minimum height=3em, right of = X, yshift = -5em, xshift=2.5cm]
            (Y) {$Y$};

            \draw[->, very thick] (X.north) --  +(0,3.5em) -- (XY.west) node[midway, above] {1. Model Training};
            \draw[->, very thick] (X.east) -- (Z.west) node[midway, above] {2. Representation};
            \draw[->, very thick] (Z.south) -- (Y.north) node[midway, right] {3. Annotation};
            \draw[->, very thick] let \p1=(X.south), \p2=(Y.west) in (\p2) -- (\x1,\y2) node[midway, above] {4. Update} -- (\p1) ;
         \end{tikzpicture} 
         \caption{System architecture: The model is trained (1) using the (partially annotated) dataset. Based on the trained model, the dataset is presented in latent space to the user (2), who can annotate samples (3). New annotated data trigger a new system cycle that also uses the new annotations (4).} 
         \label{fig:system}
     \end{subfigure}
     \hspace{.04\textwidth}
     \begin{subfigure}{.45\textwidth} 
         \centering 
     \begin{tikzpicture}
        \node[rectangle, draw, fill = {rgb:black,1;white,5}, minimum height=6.4em]
            (X) {$X$};
        \node[trapezium, rotate=-90, draw, minimum width = 6em, yshift = 0.7cm] 
            (Encoder) {\textbf{Encoder}};
        \node[rectangle, draw, fill = {rgb:black,1;white,5}, minimum height=4.8em, right of = Encoder, xshift = -0.4cm]
            (Z) {Z};
        \node[trapezium, rotate=90, draw, minimum width = 6em, below of = Z, xshift = 1.5cm] 
            (Decoder) {\rotatebox{180}{\textbf{Decoder}}};
        \node[rectangle, draw, fill = {rgb:black,1;white,5}, minimum height=6.4em, right of = Decoder, xshift = -0.4cm]
            (X2) {$\tilde{X}$};
        \node[xshift = 2.6cm, yshift = -0.1cm]{Classifier};
        \node[rectangle, draw, fill = {rgb:black,1;white,5}, minimum height=5em, xshift = 4cm, yshift = -1cm]
            (Y) {$\Tilde{Y}$};

        \readlist\Nnod{2,3} 
          \foreachitem \N \in \Nnod{ 
            \foreach \i [evaluate={\x=\Ncnt/1.2+1.4; \y=\N/4-\i/2-0.8; \prev=int(\Ncnt-1);}] in {1,...,\N}{ 
              \node[mynode] (N\Ncnt-\i) at (\x,\y) {};
              \ifnum\Ncnt>1 
                \foreach \j in {1,...,\Nnod[\prev]}{ 
                  \draw[thick] (N\prev-\j) -- (N\Ncnt-\i); 
                }
              \fi 
            }
          }

        \draw[->, very thick] (Z.east) -- +(0.2,0) |- (Decoder.north);
        \draw[->, very thick] (Z.east) -- +(0.2,0) -- +(0.2,-1.05) -- +(0.48,-1.05);
        \draw[->, very thick] let \p1=(Y.west) in (\x1-10,\y1) -- (\p1);
    \end{tikzpicture}

    \caption{Model architecture: The raw dataset is processed and represented in latent space (encoder). For training purposes the representation is reconstructed (decoder) and annotated data is presented to the classification header (classifier).}
    \label{fig:model}
     \end{subfigure} 
     \caption{Presentation of system and model architecture with $X$ (dataset), $Z$ (representation in latent space), $\Tilde{X}$ (reconstruction), $Y$ (label annotated by user) and $\Tilde{Y}$ (label predicted by classifier).} 
     \label{fig:process}
\end{figure}

%% file: content/03-Experiments.tex
\section{Experiments}

\input{figures/figure_viz_Z}

To analyze the potential of the proposed method for designing the user interface by varying hyperparameters of the \gls{ml} system, we analyzed their impact on the user experience.
As an apparatus, we created a convolutional \gls{vae} similar to the one in \cite{burgess2018understanding} with a 2D latent space and trained it on 10\,\% of the MNIST dataset \cite{MNIST}.
Data preprocessing included one hot encoding the labels, which allowed using a cross entropy loss in the classifier.
We used the common Adam optimizer with a learning rate of $5 \cdot 10^{-3}$.
Our \gls{mlp} classifier head has 10 units per layer and was trained with 100\,\% classified data.
We conducted experiments by varying the following hyperparameters:

\paragraph{\textbf{Hidden layers of classification head.}}
We hypothesized that classification heads with lower capacity would constrain the learned representation to simpler structures and thus force easier separation for the user.
To test that, we pretrained the \gls{vae} using unsupervised learning (\cref{fig:viz_Z}a, left) and added a classifier head with either two (\cref{fig:viz_Z}a, right) or none (reducing to a logistic regression, \cref{fig:viz_Z}a, center) hidden layers.
The results show that self-supervised training already leads to visible clustering.
Adding a classification head with logistic regression leads to sharper separation of classes, whereas this is not clearly observed with a higher capacity classification head.

\paragraph{\textbf{Number of training epochs.}}
For investigating the model stability, after self-supervised pretraining (i.e., the \gls{vae} component), we added the logistic regression classification head and trained it for different numbers of epochs (\cref{fig:viz_Z}b). The results show a stable change that can be continuously displayed in a tool to show changes to the user immediately.

\paragraph{\textbf{Components of the loss function.}}
We varied $\beta_{KL}$ and $\beta_{\text{classifier}}$ in \cref{eq:total_loss}.
The representation learned with neutral values is shown in the \cref{fig:viz_Z}c, left.
\Cref{fig:viz_Z}c, right and \cref{fig:viz_Z}a, center show that increasing $\beta_{\text{classifier}}$ leads to learning representations with tighter clusters.
Increasing $\beta_{KL}$ reduces the overall spread of data points (\cref{fig:viz_Z}c, center).
This is explained by the choice of prior $p(Z)$ to be a standard Gaussian for each of the dimensions of Z, and the fact that $\beta_{KL}$ scales the divergence between input representation in latent space and the chosen prior.
By choosing even higher values for $\beta_{KL}$, the second term of \cref{eq:total_loss} gains priority, which leads to a collapse where the encoder outputs $\mu_Z=0$ and $\Sigma_Z=1$ for all inputs.
The likelihood of this collapse occurring can be controlled by weighing the KL component \cite{burgess2018understanding}.

%% file: figures/figure_viz_Z.tex
\begin{figure}
    \centering
    \includegraphics[width=\textwidth]{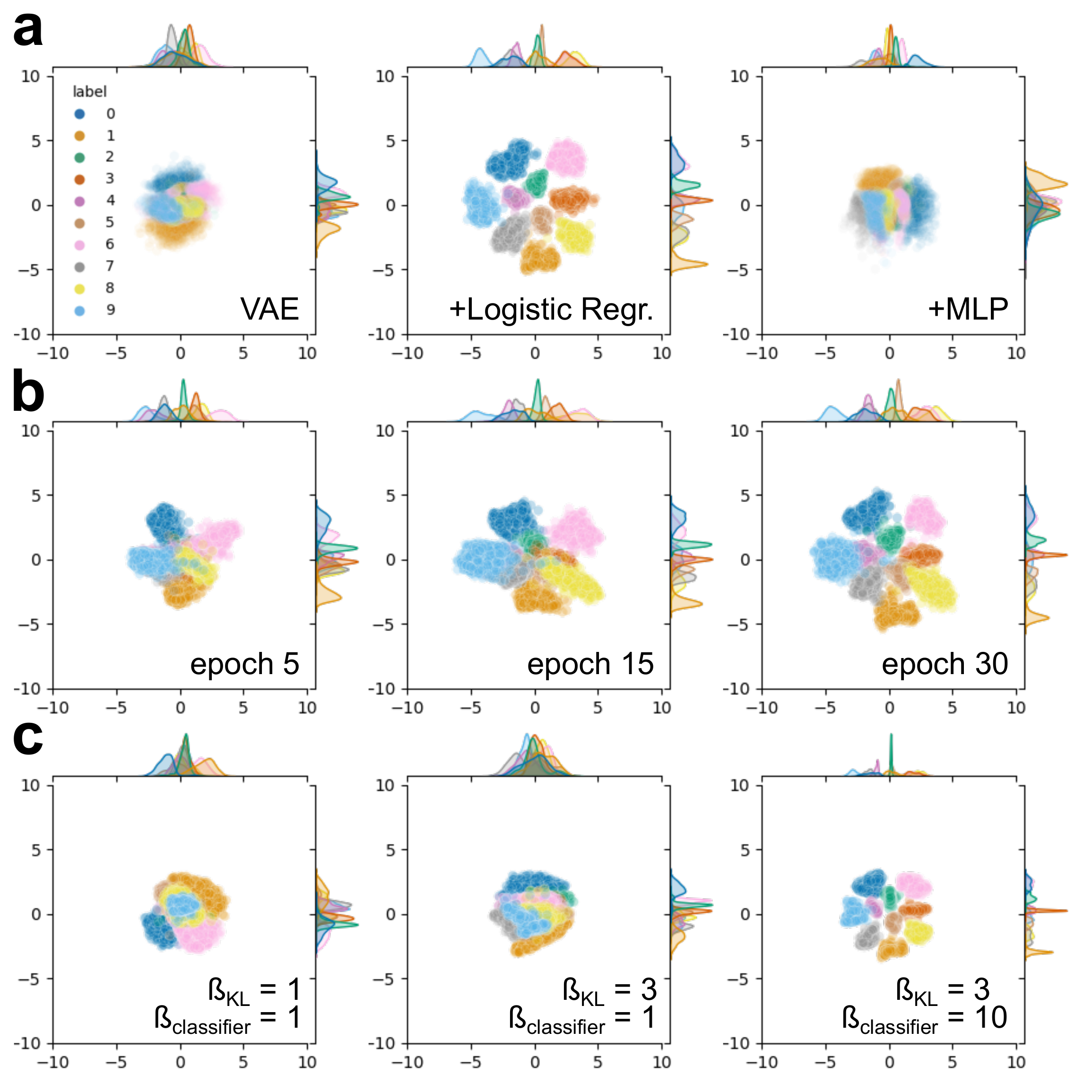}
    \caption{
        \textbf{Visualisation of MNIST dataset embedded in 2D latent space learned by model variants.}
        Scatter plots show mean of latent variable $Z$ for each input $X$; marginal plots are marginal histograms; colours denote label class.
        \textbf{(a)} Different classifier heads.
        \textit{left:} No classifier; initial condition for all other learned representations.
        \textit{center:} Logistic regression.
        \textit{right:} \gls{mlp} with 2 hidden layers.
        Hyperparameters set to $\beta_{KL} = 3$, $\beta_{\text{classifier}} = 100$, $n_{\text{epochs}}=50$.
        Logistic regression used for all other experiments.
        \textbf{(b)} Learning observed after 5, 15, and 30 training epochs; compare with initial (a, left) and final (a, center) states.
        \textbf{(c)} Different weights applied to loss components.
        Compare with
        (a, center).
    }
    \label{fig:viz_Z}
\end{figure}

%% file: content/04-Conclusion.tex
\section{Conclusions and future work}

A real need has been identified for \glsentrylong{iml}, shifting the focus from developing more accurate algorithms to improving applicability, which may include other metrics such as productivity and interpretability \cite{amershi2014power,simard2017machine}.
We have described a neural architecture that can drive a \gls{gui} for interactive representation learning for \gls{ml}-assisted data annotation, and we have found that the hyperparameters (classification header, number of training epochs, and loss function weights) have potential effects on user experience. Our network enables \gls{dm} of data in latent space by establishing three analogies relevant to an immersive experience: (latent as interface) space, (training as frame) time, and (stochastic gradient descent as physics engine) force.
We are currently exploring different learning rates to produce different degrees of stability of representations, as well as continuous learning methods \cite{hadsell2020embracing} to handle the incremental availability of labels provided by the interface front-end.